\documentclass[runningheads]{llncs}

\usepackage[utf8]{inputenc} 
\usepackage[T1]{fontenc}    
\usepackage[colorlinks]{hyperref}       
\usepackage{url}            
\usepackage{booktabs}       
\usepackage{amsfonts}       
\usepackage{nicefrac}       
\usepackage{microtype}      
\usepackage{lipsum}
\usepackage{graphicx}
\usepackage{caption}
\usepackage{subcaption}
\usepackage{cite}
\usepackage{amsmath}
\usepackage{multirow}
\usepackage{arydshln}
\usepackage{verbatim}
\usepackage{xcolor}

\makeatletter
\renewcommand*{\thetable}{\arabic{table}}
\renewcommand*{\thefigure}{\arabic{figure}}
\let\c@table\c@figure
\makeatother

\title{Generative Deep Learning Techniques for Password Generation}

\author{%
David Biesner\thanks{Equal contribution}\inst{1, 2, 3} \and
Kostadin Cvejoski$^\star$ \inst{1, 3} \and
Bogdan Georgiev\inst{1, 3} \and \\
Rafet Sifa\inst{1, 2, 3} \and
Erik Krupicka\inst{4}
}%
\institute{
Fraunhofer IAIS,\\
University of Bonn,\\
Competence Center for Machine Learning Rhine-Ruhr (ML2R),\\
Federal Criminal Police Office, Wiesbaden, Germany
}

\begin{document}
\maketitle

\begin{abstract}
    Password guessing approaches via deep learning have recently been investigated with significant breakthroughs in 
    their ability to generate novel, realistic password candidates.
    In the present work we study a broad collection of deep learning and probabilistic based models in the light of password guessing: 
    \textit{attention-based deep neural networks}, \textit{autoencoding mechanisms} and \textit{generative adversarial networks}. 
    We provide novel generative deep-learning models in terms of variational autoencoders exhibiting state-of-art sampling performance,
    yielding additional latent-space features such as interpolations and targeted sampling.
    Lastly, we perform a thorough empirical analysis in a unified controlled framework over well-known datasets (RockYou, LinkedIn, Youku, Zomato, Pwnd). 
    Our results not only identify the most promising schemes driven by deep neural networks, but also illustrate the strengths of each approach in terms of generation variability and sample uniqueness.
\end{abstract}


\section{Introduction and Motivation}

Most authentication methods commonly used today rely on users setting custom passwords to access their accounts and devices.
Password-based authentications are popular due to their ease of use, ease of implementation and the established familiarity of users and developers with the method. \cite{troyhuntblog}

However studies show that users tend to set their individual passwords predictably,
favoring short strings, names, birth dates and reusing passwords across sites. \cite{password_security_analysis, password_strength_analysis}
Since chosen passwords exhibit certain patterns and structure, it begs the question whether it is possible to simulate these patterns and generate passwords that a human user realistically might have chosen.

Password guessing is an active field of study, until recently dominated by statistical analysis of password leaks and construction of corresponding generation algorithms (see Section \ref{sec:related_work}).
These methods rely on expert knowledge and analysis of various password leaks from multiple sources to generate rules and algorithms for efficient exploitation of learned patterns.

On the other hand, in recent years major advances in machine-driven text generation have been made, notably by novel deep-learning based architectures and efficient training strategies for large amounts of training text data.
These methods are purely data driven, meaning they learn only from the structure of the input training text, without any external knowledge on the domain or structure of the data.
Major advancements in the field have been fueled by the development in several central directions such as:

\begin{enumerate}
	\item \textbf{The \textit{attention} mechanisms}. Considering a token (word, letter, sentence) within a textual environment, the idea is to develop a flexible notion of \textit{context} that connects the given token with other pieces of the textual environment. Intuitively, this allows the learning model to better grasp the textual structure (e.g. grammar, semantic meaning, word structure, etc), thus leading to eventual improvements in terms of text classification/generation/interpretability. Among others, well-known attention-based examples are given by BERT, ELMO, GPT and various further types of Transformers.
	
	\item \textbf{Model architectures and representation capabilities}. Remarkable  \\progress has been made in designing more flexible deep learning structures (CNN-based ResNets and Wide-Resnets, recurrent NNs, adversarial models, etc). The success of these deep neural networks can to a large extent be attributed to their \textit{representation capability}, i.e. they create an appropriate transformation of the data (e.g. compression or a sort of "semantic meaning" extraction) that renders the data easier to handle and solve a given problem. In this regard a central class of deep learning models is given by the so-called \textit{\textbf{autoencoders}} whose goal is not only to create a meaningful and useful data representation/transformation (Encoding) but also to be able to go back and reconstruct the initial data from the representation (Decoding). An upshot is that one could \textit{generate new data} by sampling points in the representation space and then decoding back.
	
	\item \textbf{Advanced training procedures}. The above tools would not be as efficient, had not it been for the corresponding methods to select (train) the parameters and weights of the neural networks. Among others these include appropriate momentum and annealing-driven stochastic gradient descents, Wasserstein regularization and variational approaches.
\end{enumerate}

In this paper we will continue the exploration of data driven deep-learning text generation methods for the task of password-guessing.
While some applications to password guessing already show promising results, most frameworks still can not reach or surpass state-of-the-art password generation algorithms.
Ideally, one would attempt to design more efficient password-guessing models aided by neural networks and cutting-edge practices. 
Our findings and contributions can be summarized as follows:
\begin{enumerate}
    \item We provide extensive unified analysis of previous as well as novel password guessing models based on deep learning and probabilistic techniques.
    \item The collection of architectures based on deep learning exhibits varying performance, with the top-performing models being able to reach sophisticated password generation algorithms in the password recovery task.
    \item We show that attention-driven text generation methods (Transformers) can be applied to password guessing with little additional adjustments. 
    We additionally analyse the effect of model pre-training on general language data for the password generation task against training on pure password data.
    
    \item Our novel variational autoencoder (VAE) approach allows more flexible latent representations and outperforms previous autoencoding methods based on Wasserstein training \cite{Pasquini2019}. Moreover, the VAE's performance can be compared to an attention-driven one.
    \item The VAE provides a state-of-art password matching performance as well as further sampling possibilities (interpolations, conditional and targeted sampling). However, the password latent space geometry is quite sensitive to training and regularization yielding promising grounds for future investigations in terms of conditional sampling.
\end{enumerate}

\section{Related work}
\label{sec:related_work}
Password generation has a long history outside of deep-learning architectures.
There are tools available for purely rule-based approaches (Hashcat \cite{hashcat} and JohnTheRipper \cite{JTR}), which
generate password candidates either by brute-force or dictionary attacks, 
in which a dictionary of words or previously known passwords is augmented by a set of rules,
either hand-written or machine generated \cite{gen2-rules}.

Machine-learning based approaches to password guessing may come in their most simple form as regular $n$-gram Markov Models \cite{markovmodels} or more sophisticated approaches like \emph{probabilistic context free grammar} (PCFG) \cite{PCFG},
which analyses likely structures in a password training set and applies various generation algorithms based on these observations.

Neural network based password generation has become an active field of study in the recent years.
Ranging from relatively simple recurrent neural net (RNN) architectures \cite{FLA} to recent seminal  works applying state-of-the-art text generation methods to password generation:
Generative adversarial networks (GANs) \cite{Hitaj2017, Pasquini2019}, Wasserstein Autoencoders \cite{Pasquini2019}, and bidirectional RNNs trained with the aid of pre-trained Transformer models \cite{Hang2019}.

Our work extends this palette of deep learning architectures with the Variational Autoencoder \cite{kingma2013auto} and
Transformer-based language models \cite{Radford2019}.
We additionally offer an extensive, unified and controlled comparison between the both various deep-learning based methods and more established methods mentioned above.
This analysis yields a stable benchmark for the introduction of novel models.



\section{Models}

\subsection{GAN}
A central idea of adversarial methods is the construction of generative models by game-theoretic means: a "generator" neural network produces data samples, whereas a "discriminator" neural network simultaneously attempts to discern between the real and artificially produced (by the generator) samples. The training of such a system consists in optimizing the performance of both the generator and discriminator (usually, via types of suitably chosen gradient descents and additional regularization). An important tool that smooths out gradients and makes the model more robust is the Wasserstein distance and corresponding cost function: it provides means to efficiently compute discrepancies between two given distributions $\mathbb{P}_0, \mathbb{P}_1$ as:
\begin{equation}
    W(\mathbb{P}_0, \mathbb{P}_1) = \sup_{f \in \mathcal{L}_1} \mathbb{E}_{x \sim \mathbb{P}_0} \left[ f(x) \right] - \mathbb{E}_{x \sim \mathbb{P}_1} \left[ f(x) \right],
\end{equation}
where $\mathcal{L}_1$ denotes the space of 1-Lipschitz functions. We refer to \cite{Goodfellow2014}, \cite{Arjovsky2017}, \cite{Gulrajani2017} for further background.

Concerning password guessing and generation our starting point is the well-known PassGAN model proposed in \cite{Hitaj2017}. A further substantial breakthrough in this direction of GAN-based models was given in \cite{Pasquini2019}. The original PassGAN defines a discriminator and generator in terms of residual networks \cite{Zagoruyko2016} - these are assembled from the so-called residual blocks (e.g. a stack of convolutional neural networks followed by a batch-normalization \cite{Ioffe2015}). A specific feature here is that the input partially bypasses the residual block (shortcutting) and is added to the output of the residual block - the aim is to diminish the effect of vanishing gradients and introduce a form of "memory" across the residual network.

\subsection{Representation Learning with Deep Latent Variable Models}
Many real world tasks as well as machine learning tasks can be solved very difficult or very easy depending on how the information is represented. Let us take for example the task of dividing the number 180 by 2. This task is straight forward using the Arabic numeral system. Now, we pose the same problem but we use the Roman numeral system to represent the numbers. In this case the task is dividing CLXXX by II. Using the Roman numeral system the task seems much more difficult for a modern educated person \cite{Goodfellow-et-al-2016}.

Representation learning is interesting because it provides one particular way to perform unsupervised learning. Unsupervised deep learning algorithms have a main training objective but also learn representations as side product. Naturally arises the question: \textit{What makes one representation better than another?} In the context of machine learning a representation $\mathbf{z}$ is good if it makes the subsequent task easy to solve. One hypothesis out there is that the best representation is the one in which each of the features within the representation corresponds to the underlying factors or causes that generated the observed data. In ideal case each of the features or directions in the features space corresponds to different causes of the data. This kind of representation disentangles the underlying generative factors and they are called \textit{disentangled} representation. There is a large body of deep representation learning research focused on obtaining disentangled representation.

\textit{Why representation learning is important in password guessing setting?} There has been extensive research on text-password attacks and how people are choosing passwords. Human chosen passwords are not uniformly distributed in the space of passwords (all possible strings). The users tend to pick passwords that are easy to remember or have some personal meaning. Also, every web services has different policies about password construction (length, number of symbol, capital letters and etc.) Therefore, one can conclude that there are different latent factors in constructing a password. If we are able to learn these different factors, it will be possible to generate passwords that are from the same distribution as the human generated passwords.

\subsubsection{Variational Auto Encoder (VAE)}
\label{ssec:vae}
\cite{kingma2013auto} is a framework for efficient optimization of \textit{deep latent variable models} (DLVM). It comprises of two main components: i) \textit{encoder} and ii) \textit{decoder}. The encoder is stochastic function $\phi: \mathbf{X} \rightarrow\mathbf{Z}$ that maps the input space (passwords) $\mathbf{X}$ to the latent space $\mathbf{Z}$. The decoder is deterministic function that maps a code from the latent space to the input space $\theta: \mathbf{Z} \rightarrow\mathbf{X}$. The model is trained by maximizing the log likelihood

\begin{equation}
  \mathcal{L}(\mathbf{\theta, \phi, x^{(i)}})=-D_{KL}(q_{\phi}(\mathbf{z}|\mathbf{x}^{(i)})||p_{\theta}(\mathbf{z}))+\mathbb{E}_{q_{\phi}(\mathbf{z}|\mathbf{x}^{(i)})}\left[\log{p_{\theta}(\mathbf{x}^{(i)}|\mathbf{z})}\right].
\label{eq:vae_loss}
\end{equation}

The model learns to reconstruct the password $\mathbf{x}$ given to the input by first, mapping the password to a distribution of latent codes $p_{\psi}(\mathbf{z}|\mathbf{x})$ then we sample from the posterior distribution and we pass the latent code $\mathbf{z}$ to the decoder $p_{\theta}(\mathbf{x}^{(i)}|\mathbf{z})$. During training a strong prior $p(\mathbf{z})$ is imposed on the learned latent code distribution. Setting the prior can be informed by some previous knowledge that we have about the generative process. However, usually in the VAE framework the prior is set to be centered isotropic Gaussian distribution $p(\mathbf{z})=\mathcal{N(\mathbf{z}; \mathbf{0}, \mathbf{I})}$. this is done so later we can easy sample from them and generate new passwords.

The latent space learned by the encoder imposes a geometric connections among latent points that have some semantic similarity in the data space. As a result \textit{similar} points in the data space have latent representation that are close to each other. The notation of \textit{similarity} depends on the modeled data, in the case of password generation it can be based on the structure of the password, the common substring, etc.

Training of VAEs with unmodified objective function \eqref{eq:vae_loss} often can lead to converging to a undesirable local minimum \cite{bowman2015generating}, \cite{sonderby2016train}, \cite{kingma2019introduction}. This is the case because at the start of the training, the reconstruction term $\log{p_{\theta}(\mathbf{x}|\mathbf{z})}$ is weak, i.e., the latent code $\mathbf{z}$ does not contain any useful information about the point $\mathbf{x}$. This will lead to an optimization surface with a local minimum around $q(\mathbf{z}|\mathbf{x})\approx{p(\mathbf{z})}$, which is difficult to escape. One possible solution (\cite{bowman2015generating}, \cite{sonderby2016train}) to this problem is to use an optimization scheduler where the weight $\beta$ for the $D_{KL}(q_{\phi}(\mathbf{z}|\mathbf{x})||p_{\theta}(\mathbf{z}))$ is annealed for 0 to 1. 

\subsubsection{Wasserstein Auto Encoder (WAE)}
\label{ssec:wsae}
\cite{tolstikhin2017wasserstein} is a framework for building generative models using the optimal transport theory (OT) and by minimizing the optimal transport cost \cite{villani2003topics} between the data (unknown) distribution $P_X$ and a latent variable model $P_G$. The optimal transport cost measures the distance between two probabilities by posing much weaker structure in comparison to other distance metrics, like for example $f$-divergence. This is important when we deal with data that is supported on low dimensional manifold in the input space. The WAE objective is defined as
\begin{equation}
    \label{eq:wae-objective}
    D_{WAE}(P_X,P_G):= \inf_{Q(Z|X)\in\mathcal{Q}}\mathbb{E}_{P_X}\mathbb{E}_{Q(Z|X)}\left[c\left(X, G(Z)\right)\right] + \lambda \mathcal{D}_Z(Q_Z, P_Z),
\end{equation}

where $\mathcal{Q}$ is any nonparametric set of probabilistic encoders, $\mathcal{D}_Z$ is an arbitrary divergence between $Q_Z$ and $P_Z$, and $\lambda>0$ is hyper-parameter. 

\subsection{Transformers}
\label{ssec:transformers}
In recent years the transformer,
originally applied to machine translation in \cite{Vaswani2017},
have become increasingly popular,
with transformer-based architectures setting new benchmarks in text generation, machine translation and other NLP tasks.

Transformers rely almost solely on self-attention to process an input text,
which considers all pairs of words in the sentence instead of the linear sequence of words.
While RNNs may lose the memory of words in the beginning of a sentence rather quickly,
transformers are able to capture long-term dependencies between words in a sentence and between sentences.

The self-attention mechanism evaluates attention for each word pair by multiplying their entries in the query, key and value matrices $Q, K \in  \mathbb{R}^{n, d_k}$ and $V \in \mathbb{R}^{n, d_v}$ ($d_k$ dimensionality of the query and key, $d_v$ of the value vectors respectively) as
\begin{equation}\label{eq:attention}
    \textbf{Attention}(Q, K, V) = \textbf{softmax}\left(\frac{QK^T}{\sqrt{d_k}}\right)V
\end{equation}
where the output of the softmax operation provides for each word a probability distribution over all other words in the sequence,
which is then used to weight the word values to produce the attention output.

Popular transformer architectures include BERT \cite{Devlin2019}, XLNet \cite{XLNet}, Transformer XL \cite{TransformerXL}, and GPT2 \cite{Radford2019}. 
In our work we apply the GPT2 architecture to the password modeling task. 

GPT2 is a transformer-based language model, trained on the causal language modeling (CLM) objective, meaning given an incomplete sentence it will try to predict the next upcoming word. 
A tokenized input sentence is therefore read by a transformer block which outputs a probability distribution $p_\theta$ over the vocabulary.
For a corpus of tokens $\mathcal{U} = (u_1, \dots, u_N)$ and a context window $k\in\mathbb{N}$ the loss is then given as
\begin{equation}\label{eq:gpt2_loss}
    \mathcal{L}(\mathcal{U}) = \sum_{i=1}^N p_\theta(u_i | u_{i-1}, \dots, u_{i-k})
\end{equation}

To generate text, given a text prompt \mbox{(e.g. ``Hello my '')} the model will start generating text that continues the sentence \mbox{(``name is GPT2!'')}.
We provide details on the model and training in Section \ref{sec:training_gpt2}.

\section{Data}
\subsection{On public datasets}
There are several datasets of passwords publicly available.
These lists contain passwords that were at some point in time leaked to the public from on certain websites.
Leaks contain in rare cases plaintext passwords (e.g. the RockYou leak), but more commonly only password hashes that are then recovered using password guessing methods.
Password datasets contain either passwords from a single leak or are aggregated from several leaked sources.


The specific password datasets\footnote{Provided link references accessed on 7.12.2020} we employ for training or evaluation are:
\begin{itemize}
    \item `rockyou'\footnote{\url{https://www.kaggle.com/wjburns/common-password-list-rockyoutxt}}: 13.0M passwords (116MB), leak of plaintext passwords from a single source,
    \item `Have I Been Pwnd V1' (`pwnd')\footnote{\url{https://hashes.org/leaks.php?id=70}}: 319.8M passwords (3.2GB), compilation of multiple leaks of hashed passwords, almost all passwords are recovered and available as plaintext,
    \item `linkedin'\footnote{\url{https://hashes.org/leaks.php?id=68}}: 60.1M passwords (619MB), leak of hashed passwords from a single source, most passwords are recovered and available as plaintext,
    \item Addtional leaks for evaluation: `myspace' (53k), `yahoo' (430k), `youku' (48M), `seclist' (969k), `skullsecurity' (6.2M), `zomato' (6.3M)\footnote{
    \url{https://weakpass.com/wordlist/22}, 
    \url{https://weakpass.com/wordlist/44}, 
    \url{https://hashes.org/leaks.php?id=508},  
    \url{https://hashes.org/leaks.php?id=587},  
    \url{https://weakpass.com/wordlist/50}, 
    \url{https://weakpass.com/wordlist/671}, 
    }
\end{itemize}

Note that there is a certain bias inherent in password datasets.
Passwords are (or should be) generally not stored as plaintext but as hashes.
Password guessing methods however need plaintext passwords for training.
Therefore all available data either belongs to websites which stored their passwords as plaintext (e.g. rockyou)
or are recovered password hashes, 
therefore passwords that were already possible to reconstruct using available methods.

\subsection{On segments -- splitting passwords into likely subword tokens}
\label{sec:BPE}

Classically, text generation models are based on words or characters.
Either a model chose at each step a new likely word from a large vocabulary (eg. $300k$) to generate,
or it constructed words character by character, choosing from all available characters (for English text usually around 75 characters, a-z, A-Z, 0-9 and punctuation).

While the first approach is not suitable for password generation, 
since we only aim to generate one "word" (i.e. password) at a time and would need a vocabulary consisting of all possible passwords in the first place,
the character-based approach is an immediate fit to our task.
By pre-processing our dataset to only contain passwords consisting of the 75 characters we allow (depending on the dataset this removes 0-10\% of all entries) we can tokenize each password into a sequence of characters and generate every possible password within the confines of our vocabulary.

However, one could ask whether it is really necessary for the model to take 8 generation steps to generate the sequence \texttt{password},
which occurs very regularly in the dataset.
It might be beneficial to be able to generate certain common strings (\texttt{password}, \texttt{123}, \texttt{love}) in one step 
to decrease the length of common sequences since many sequence generation models struggle with remembering context over long distances \cite{LSTM}.

Recent NLP models therefore employ a method called byte-pair-encoding \cite{BPENeuralMachineTranslation},
which searches a text corpus for the $N$ most common sequences of characters, builds a vocabulary of size $N$
and tokenizes new text using this vocabulary.
The vocabulary size $N$ is usually in the range of $30k$, far greater than character based models but smaller than word based models.
Given that all relevant single characters are part of the vocabulary this method is then able to tokenize (and therefore encode and generate) every possible string.

In our work we employ this method as follows:
\begin{enumerate}
    \item Choose training password dataset (rockyou)
    \item Split each password into pure segments (ie. containing only characters or only numbers or only punctuation): \\
    \texttt{pass\_word!!123} $\rightarrow$ \texttt{pass \_ word !! 123}
    \item Count all segments and sort by frequency
    \item Choose most common $N=30k$ segments, additionally add all single characters that are not yet in the vocabulary
\end{enumerate}

We end up with a vocabulary of around $30k$ pure segments that are common substrings in the dataset. 
Using this vocabulary we are then able to tokenize every password in this or other datasets concisely.
Further work might include using the exact byte-pair-encoding method described in \cite{BPENeuralMachineTranslation} to
extract a vocabulary of statistically significant non-pure substrings.

\section{Results}

\subsection{Experimental Setup}

\subsubsection{GAN-based models}
In our setup, we essentially utilize the PassGAN as a standard benchmark - this is well motivated by the previous substantial studies of GAN-based models (\cite{Hitaj2017}, \cite{Pasquini2019}). The generator/discriminator are defined as residual neural networks consisting of 6 standard residual blocks followed by a linear projection/softmax function, respectively. Each of the standard residual blocks consists of 2 convolutional layers (with kernel-size 3, stride 1, no dilation) followed by a batch-normalization layer. The generator's latent space (i.e. the input space) is set to 256 inspired by \cite{Hitaj2017}.

The PassGAN training was based on some state-of-art practices such as Wasserstein GAN and gradient clipping (cf. \cite{Arjovsky2017}), as well as gradient penalty regularization (cf. \cite{Gulrajani2017}). We mostly used a batch-size of 256, a gradient-penalty-hyperparameter $\lambda$ set to 10 and 10 discriminator iterations per generator step. The preferred gradient descent was based on ADAM \cite{kingma2014adam} with an initial learning rate of $\sim 10^{-4}$ and beta (momentum) parameters $0.5, 0.9$; a fixed-interval annealing with an iteration step of $\sim 10^6$ was also used.

\subsubsection{WAE/VAE-based models}
 \label{sssec:training_vae_wae}
 
 For both the encoder and the decoder we use a CNN with fixed kernel size of 3. The depth and the dilation of the convolution is gradually increase from [1, 2, 4], [1, 2, 4, 8], [1, 2, 4, 8, 16] to [1, 2, 4, 8, 16, 32]. We use cross-validation to pick the best hyper-parameters for the model. The number of channels for the CNN block is 512, the latent dimension of the model $\mathbf{z}$ is chosen from [64, 128, 256]. We use the ADAM \cite{kingma2014adam} with learning rate $\sim 10^{-4}$ and momentum $\beta_1=0.5$ and $\beta_2=0.9$. The batch size is chosen to be 128 and we also use early stopping. Following \cite{bowman2015generating}, we use KL cost annealing strategy. In the case of the WAE we use \textit{maximum mean discrepancy (MMD)} regularizer with RBF kernel.

\subsubsection{GPT2-based models}\label{sec:training_gpt2}
\begin{figure}[t]
\centering
    \begin{subfigure}[t]{0.5\textwidth}
    \centering
        \begin{quotation}
        \scriptsize{%
        \include{figures/text_obama}
        }
        \end{quotation}
    \end{subfigure}
    \hspace{0.1cm}
    \begin{subfigure}[t]{0.55\textwidth}
    \centering
        \begin{quotation}
        \scriptsize{%
        \include{figures/text_passwords}
        }
        \end{quotation}
    \end{subfigure}
    \caption{(a) Training data for the pretrained GPT2 model is raw text from various inter sources. (b) Training data for finetuning of GPT2 is passwords from the `rockyou' dataset contatenated into continuous text.}
    \label{fig:gpt2_text}
\end{figure}

The original openly available GPT2 model is trained on a large corpus of internet text.
Training data is provided simply as a sequence of raw, unlabeled text that is fed into the model.
For training on passwords, one can therefore take a dataset of passwords and construct training data by concatenating shuffled passwords into continuous text (see also Figure (\ref{fig:gpt2_text})). In our report we train two GPT2-based models:

(i) We finetune (i.e. continue the training of) the pre-trained model with our password dataset.
We hope that the original training gives the model some background on how language is generally structured as well as a vocabulary of common words.
Finetuning on passwords should then give the model an understanding of the structures and the vocabulary of passwords and force it to generate passwords when prompted.
We call this model GPT2-Finetuned or GPT2F;

(ii) The other model only uses the architecture of GPT2 and trains a randomly initialized model from scratch.
A concern using a pre-trained and finetuned model is whether the model resorts back to generating regular English text when faced with certain prompts, this problem with a model trained from scratch will not appear since all the text it has ever known are passwords.

We train the model using an openly available implementation of GPT2,\footnote{\url{https://huggingface.co/transformers/model_doc/gpt2.html}}
with the default \emph{gpt2}\footnote{\url{https://huggingface.co/transformers/pretrained_models.html}} model as pretrained base and default hyperparameters.
The model is therefore a 12 layer, 12 attention-head model with latent dimension 768, maximum sequence length of 1024 and a vocabulary size of 50257.
GPT2 applies byte-pair encoding trained on general english text to tokenize text (see Section \ref{sec:BPE}).

\subsection{Most common generated passwords}

\begin{table}[]
\centering
    \begin{subtable}[h]{\textwidth}
    \centering
    \resizebox{0.75\width}{!}{%
    \begin{tabular}{c|cc @{\hskip 1cm} c|cc @{\hskip 1cm} c|c}
    &&&&&& \\
    Model & `rockyou' & `pwnd' & Model & `rockyou' & `pwnd' & Model & rockyou\\
    \cmidrule(lr{0.5cm}){1-3}
    \cmidrule(lr{0.5cm}){4-6}
    \cmidrule(lr{0.5cm}){7-8}
    
    \multirow{10}{*}{GPT2S} 
    & \texttt{love} & \texttt{2010} &  
            \multirow{10}{*}{VAE}  
            & \texttt{leslie} & \texttt{MARIA} &  
                \multirow{10}{*}{PassGAN}  
                    & \texttt{123456} \\
    
    & \texttt{mrs.} & \texttt{love} & & \texttt{yankee} & \texttt{hilton} & & \texttt{12345} \\
    & \texttt{baby} & \texttt{1234} & & \texttt{kirsty} & \texttt{SEXY} & & \texttt{123456789} \\
    & \texttt{sexy} & \texttt{2000} & & \texttt{jeremy} & \texttt{4678} & & \texttt{1234567} \\
    & \texttt{girl} & \texttt{2009} & & \texttt{claudia} & \texttt{NATA} & & \texttt{12345678} \\
    & \texttt{angel} & \texttt{2008} & & \texttt{gangsta} & \texttt{ALEX} & & \texttt{angela} \\
    & \texttt{1992} & \texttt{12345} & & \texttt{violet} & \texttt{JOSE} & & \texttt{angels} \\
    & \texttt{1993} & \texttt{2011} & & \texttt{andrei} & \texttt{BABY} & & \texttt{angel1} \\
    & \texttt{1994} & \texttt{2007} & & \texttt{jennifer} & \texttt{MAMA} & & \texttt{buster} \\
    & \texttt{2007} & \texttt{1992} & & \texttt{natalie} & \texttt{ANGEL} & & \texttt{128456} \\
    
    \cmidrule(lr{0.5cm}){1-3}
    \cmidrule(lr{0.5cm}){4-6}

    \multirow{10}{*}{GPT2F} 
    & \texttt{ilove} & \texttt{1234} &  
        \multirow{10}{*}{WAE}  
        & \texttt{08970899} & \texttt{pearlpolina} \\
    
    & \texttt{love} & \texttt{2010} & & \texttt{08520899} & \texttt{larawilliam} \\
    & \texttt{baby} & \texttt{love} & & \texttt{08970897} & \texttt{manuchandler} \\
    & \texttt{iluv} & \texttt{2000} & & \texttt{0897} & \texttt{parispolina} \\
    & \texttt{sexy} & \texttt{2009} & & \texttt{08970852} & \texttt{pearlleft} \\
    & \texttt{pink} & \texttt{2008} & & \texttt{08990899} & \texttt{poohilove} \\
    & \texttt{memyself} & \texttt{12345} & & \texttt{0852} & \texttt{1609polina} \\
    & \texttt{caoimhe} & \texttt{2011} & & \texttt{08520897} & \texttt{larapolina} \\
    & \texttt{cintaku} & \texttt{2007} & & \texttt{54040899} & \texttt{glass8159} \\
    & \texttt{jess} & \texttt{1987} & & \texttt{08520852} & \texttt{pearlwilliam} \\
    
    \end{tabular}
    }
    \caption{Most common generated passwords per model, separated by training dataset.}
    \label{tab:most_common_models}
    \end{subtable}
    
    \begin{subtable}[h]{\textwidth}
    \centering
    \resizebox{0.75\width}{!}{%
    \begin{tabular}{rc@{\hskip 0.5cm}rc@{\hskip 0.5cm}rc@{\hskip 0.5cm}rc@{\hskip 0.5cm}rc}
    &&&&&&& \\
    \multicolumn{2}{c}{Hashcat -- best64} & \multicolumn{2}{c}{Hashcat -- gen2} & \multicolumn{2}{c}{PCFG} & \multicolumn{2}{c}{Markov Model} & \multicolumn{2}{c}{PRINCE}\\
    Freq & Password & Freq & Password & Freq & Password & Freq & Password & Freq & Password \\
    \midrule
    268 648 & \texttt{08} & 470032 & \texttt{} & 8 & \texttt{zaq12345} & 2 002 838 &  \texttt{12} & 6 & \texttt{22062206} \\
    224 033 & \texttt{ma} & 59840 & \texttt{1} & 8 & \texttt{zaq1234} & 1 697 575 &  \texttt{08} & 4 & \texttt{shie2206} \\
    125 357 & \texttt{sa} & 48516 & \texttt{2} & 8 & \texttt{zaq121} & 1 589 461 &  \texttt{ma} & 4 & \texttt{love2206} \\
    123 535 & \texttt{10} & 42536 & \texttt{0} & 8 & \texttt{aq12345} & 1 429 925 &  \texttt{01} & 4 & \texttt{june2206} \\
    118 000 & \texttt{20} & 40760 & \texttt{3} & 8 & \texttt{123wed} & 1 011 725 &  \texttt{10} & 4 & \texttt{july2206} \\
    117 682 & \texttt{01} & 33257 & \texttt{4} & 8 & \texttt{123was} &  974 664 &  \texttt{00} & 4 & \texttt{contrasea} \\
    115 167 & \texttt{ch} & 32531 & \texttt{ *} & 7 & \texttt{zaq123456} &  924 150 &  \texttt{02} & 4 & \texttt{asdfghjkl} \\
    114 823 & \texttt{ja} & 31886 & \texttt{a} & 7 & \texttt{tre4567} &  922 786 &  \texttt{09} & 4 & \texttt{22062536} \\
    109 098 & \texttt{ca} & 30455 & \texttt{5} & 7 & \texttt{tre456} &  855 534 &  \texttt{99} & 4 & \texttt{22062534} \\
    108 894 & \texttt{09} & 29998 & \texttt{9} & 7 & \texttt{sw2121} &  820 617 &  \texttt{an} & 4 & \texttt{22062533} \\
    \end{tabular}
    }
    \caption{Most common generated passwords per comparison method trained on `rockyou' with corresponding frequency for $10^9$ generated passwords. Most common password generated by Hashcat gen2 is an empty string.}
    \label{tab:most_common_comparison}
    \end{subtable}
    
    \begin{subtable}[h]{\textwidth}
        \centering
    \resizebox{0.75\width}{!}{%
        \begin{tabular}{l|r}
        & \\
             Model  & Relative Frequency \\
             \midrule
            PassGAN (rockyou) & $9.9 \times 10^{-3}$ \\
            GPT2S (rockyou)    & $2.5 \times 10^{-4} $\\
            WAE (rockyou) & $1.8 \times 10^{-4}$ \\
            GPT2F (pwnd)    & $1.6 \times 10^{-4} $\\
            GPT2S (pwnd)    & $1.4 \times 10^{-4} $ \\
            GPT2F (rockyou)     & $1.6 \times 10^{-5} $\\
            VAE (rockyou) & $1.2 \times 10^{-5}$ \\
            WAE (pwnd) & $5.4 \times 10^{-7}$ \\
            VAE (pwnd) & $4.6 \times 10^{-7}$ \\
        \end{tabular}
        }
        \caption{Relative frequency of most common generated password per model. Sorted by frequency, lower models generate the most common password less often.}
        \label{tab:most_common_frequencies}
    \end{subtable}
    
    \caption{Analysis of most common generated passwords of our models trained on `rockyou' and pwnd, as well as the comparison methods evaluated in Table (\ref{tab:linkedin_comparison}).}
    \label{tab:most_common}
\end{table}

In Table (\ref{tab:most_common}) we compare the most common passwords generated by our models along with the most commonly generated passwords by the comparison methods. 
We observe in Table (\ref{tab:most_common_models}) that each model and training dataset generates a unique set of password candidates.
While the GPT2-based models trained on `pwnd' commonly generate year numbers and short sequences of digits,
the equivalent models trained on `rockyou' produce strings that generally look more like real words that one might commonly find as a password or password substring.
The VAE model focuses for both training datasets on names, with the `rockyou' model generating lowercase names and the `pwnd' model prefering all-caps strings.
For each of these mentioned models we observe that the model trained on `pwnd' in general seems to produce shorter strings.

However, the WAE reverses the behaviour observed so far.
The `rockyou' version generates an exceeding amount of similar looking numbers, while the pwnd-based model generates all lowercase strings that mostly appear like a concatenation of real words. 

Finally we consider the PassGAN model implemented on the `rockyou' training data.
Here we observe a combination of simple number strings containing some variation of \texttt{12345} and character strings based on the substring \texttt{angel}.
In general the most common passwords seem similar to each other, with less variance as observed for the other models.

Table (\ref{tab:most_common_frequencies}) describes the relative frequency of the most common generated password candidate for our models.
We again see a large variance between the models and training data basis.
While the PassGAN produces its most common string (\texttt{123456}) more every 10000 generations,
the VAE trained on `pwnd' produces its most common string (\texttt{MARIA}) only once every 2M generated passwords.

In general both VAE models generate their most common passwords comparatively rarely,
while the GPT2-based models all appear in the middle of the field.

\begin{figure}[t]
     \centering
     \begin{subfigure}[t]{0.7\textwidth}
         \centering
         \includegraphics[width=\textwidth]{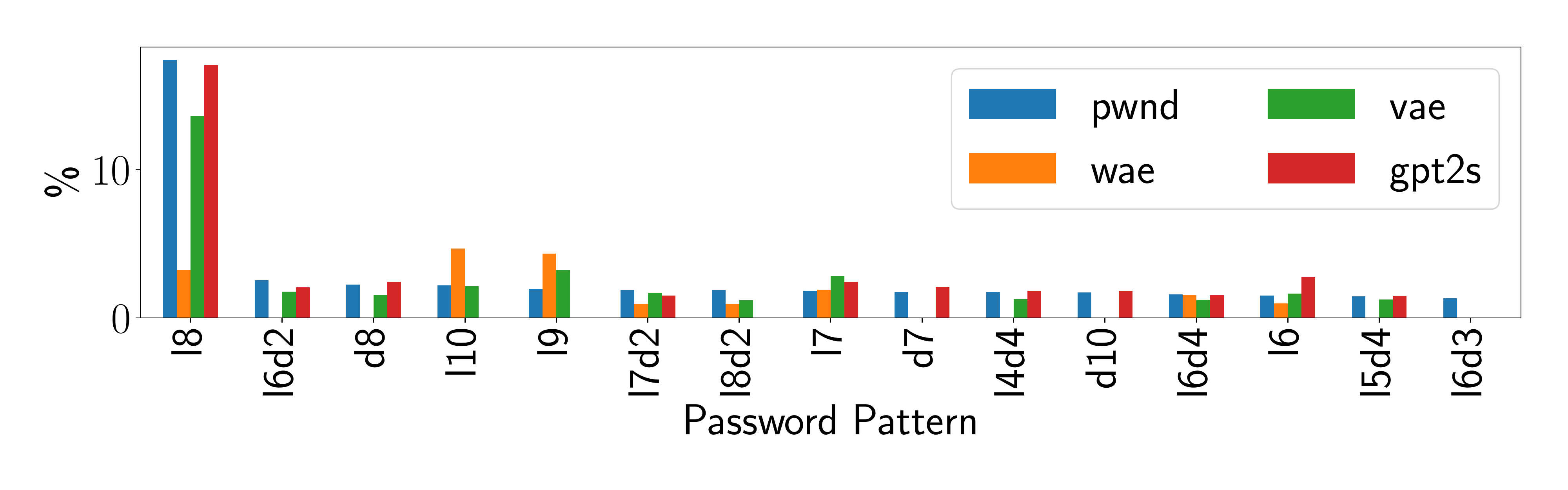}
         \caption{Histogram for different password patterns.}
     \label{fig:pwnd_pass_pattern_stats}
     \end{subfigure}
      \begin{subfigure}[t]{\textwidth}
         \centering
         \includegraphics[width=0.7\textwidth]{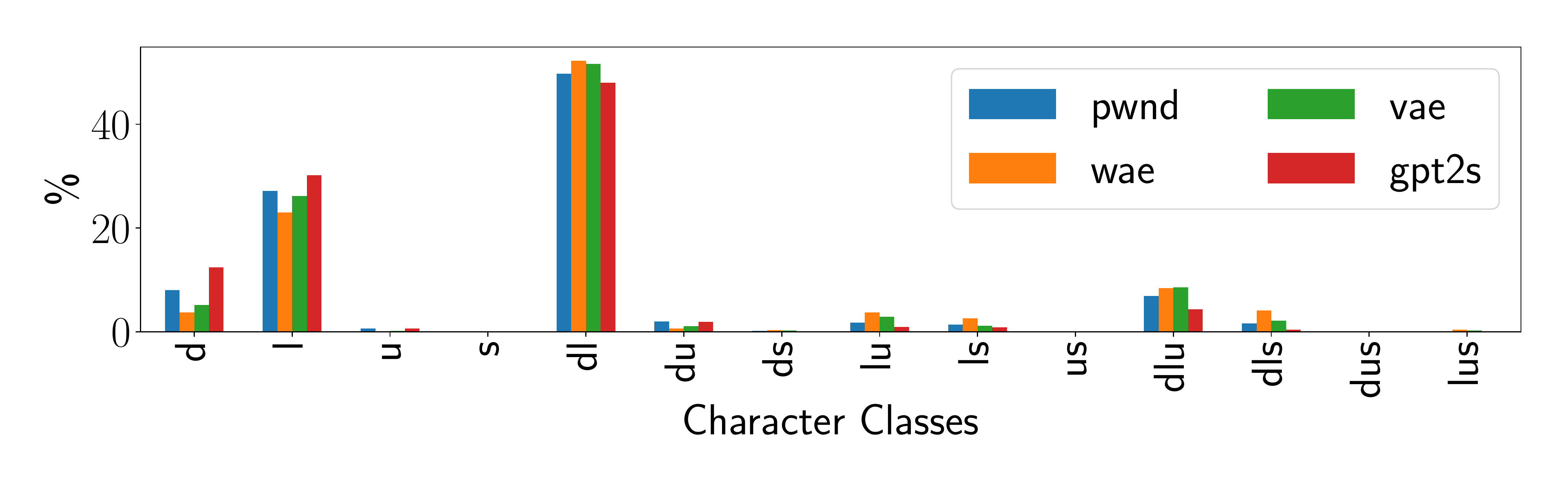}
         \caption{Histogram for different character classes.}
     \label{fig:pwnd_pass_char_classes_stats}
     \end{subfigure}
     \begin{subfigure}[t]{\textwidth}
         \centering
         \includegraphics[width=0.7\textwidth]{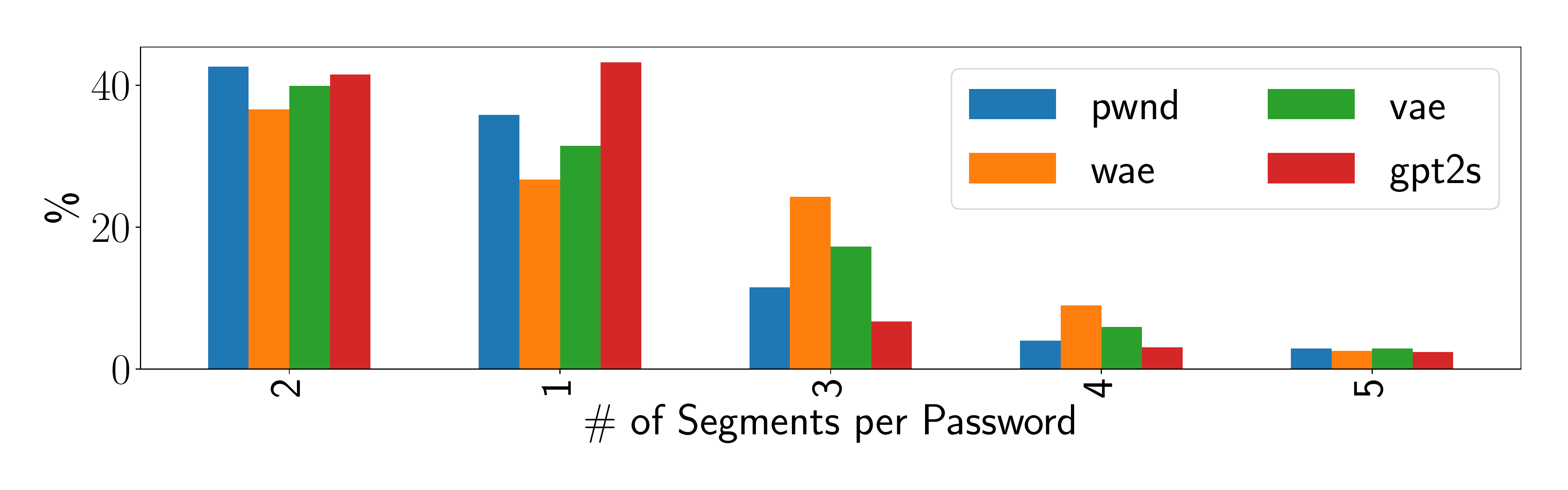}
         \caption{Histogram for the number of segments per password.}
     \label{fig:pwnd_pass_segments_stats}
     \end{subfigure}
     \caption{
     We compare the properties of passwords generated using models trained on the `pwnd' dataset against the properties of the passwords contained in the `pwnd' dataset.}
     \label{fig:pwnd_models_stats}
\end{figure}

 Additionally, to estimate if our trained model follow the distribution of the empirical dataset (the training set), we introduce three different statistics:
 \begin{itemize}
     \item \textit{PCFG-like} \cite{PCFG} - password \texttt{password!23} is converted to \textbf{L8S1D2} where \textbf{L} is alpha character, \textbf{S} is special symbol character and \textbf{D} is a digit; 
    
    \item \textit{character type composition} - is a representation of the password with respect to character types contained in the password (ex. \texttt{Password!23} is represented as \textbf{lusd}) where \textbf{l} is a lower case letter, \textbf{u} is an upper case letter, \textbf{s} is a special symbol and \textbf{d} is digit;
    \item the number fragments per password (see Section \ref{sec:BPE}). 
 \end{itemize}
 In Figure (\ref{fig:pwnd_models_stats}) are presented the three statistics on passwords generated from models trained on the `pwnd' dataset. The most common password structure for the ´pwnd' dataset is \textbf{l8} (see Figure (\ref{fig:pwnd_pass_pattern_stats})) with almost 20\% of the whole dataset. The GPT2S model successfully matches the number of passwords with structure \textbf{l8}, and the WAE models is performing poorly. For the rest of the password structures the GPT2S and the VAE model perform roughly the same and more or less they match the underlying empirical distribution. The second statistics presented in Figure (\ref{fig:pwnd_pass_char_classes_stats}) is the distribution of the passwords with respect to the type of characters contained in the password. Half of the passwords approx. 50\% in the ´pwnd´ dataset contain digits and lowercase letters (\textbf{dl}). All of the models roughly are matching the number of passwords containing \textbf{dl}. The GPT2S model generates more passwords containing only digits than the other models including also the empirical data. The last statistics is the distribution over the number of segments per password, i.e., the length of a password. Roughly 42\% of the passwords in the `pwnd' dataset are composed of two segments. From Figure (\ref{fig:pwnd_pass_segments_stats}) one can see that GPT2S prefers generating passwords composed of one segment. Whereas, the VAE is roughly following the empirical data distribution and most of the generated passwords are with two segments. As the number of segments grows the difference between the empirical distribution and the distribution of the passwords generated by the models is reducing.

\subsection{Password guessing performance}

\begin{table}[t]
    \centering
    \begin{tabular}{l|rrrrrrrrr}
    \multicolumn{1}{l}{\textbf{Model}} 
    & \multicolumn{1}{c}{\textbf{rockyou}} & \multicolumn{1}{c}{\textbf{linkedin}}  
    & \multicolumn{1}{c}{\textbf{pwnd}} & \multicolumn{1}{c}{\textbf{myspace}} & \multicolumn{1}{c}{\textbf{yahoo}}
    & \multicolumn{1}{c}{\textbf{seclist}} & \multicolumn{1}{c}{\textbf{skullsec}}  
     & \multicolumn{1}{c}{\textbf{youku}} & \multicolumn{1}{c}{\textbf{zomato}} \\   \cline{1-10}
    VAE*         & 44.9\%  & \textbf{21.8\%} & 15.4\% & 62.5\% & 47.3\% & 57.4\% & 32.1\% & 13.8\%  & 20.3\%   \\
    WAE*         & 28.2\%  & 12.9\% & 8.6\% & 45.7\% & 34.6\% & 43.8\% & 22.3\% & 10.1\%  & 13.7\%   \\
    PassGAN*     & 15.9\%  & 6.8\% & 4.7\% & 24.6\% & 19.0\% & 30.7\% & 13.6\% & 5.1\%  & 8.3\%   \\
    GPT2F*       & \textbf{45.1\%}  & 20.3\% & 14.7\% & \textbf{65.8\%} & \textbf{47.6\%} & 55.3\% & 31.1\% & 11.8\% & 18.6\%   \\
    GPT2S*       & 41.7\%  & 18.7\% & 13.9\% & 61.1\% & 45.0\% & 53.6\% & 31.0\% & 13.4\% & 17.8\%   \\
    VAE**        & 26.7\%  & 13.5\% & 14.6\% & 44.7\% & 33.1\% & 46.2\% & 24.8\% & 9.7\%  & 16.5\%   \\
    WAE**        & 11.7\%  & 5.2\% & 3.6\% & 20.5\% & 16.7\% & 26.0\% & 11.2\% & 3.6\%  & 6.2\%   \\
    GPT2F**      & 36.4\%  & 19.9\% & 22.1\% & 57.1\% & 42.4\% & 56.5\% & 34.8\% & 14.4\% & 24.3\%   \\
    GPT2S**      & 37.6\%  & 20.7\% & \textbf{22.7\%} & 58.3\% & 43.7\% & \textbf{58.0\%} & \textbf{36.0\%} & \textbf{14.5\%} & \textbf{25.3\%}   \\
    \end{tabular}
    \caption{Evaluate the model password-matching performance on all datasets with $10^9$ generated passwords. (*) Models trained on the `rockyou' dataset; (**) Models trained on `pwnd' dataset. Evaluation is done on the full size of the dataset except for the `rockyou' and `pwnd' where the models are evaluated only on the test set (20\% of the full size).}
	\label{tab:dl-models-results}
\end{table}

To evaluate the power of our password generation methods we match generated passwords to a predefined test set.
Given a dataset of passwords we split into train and test (80\% / 20\%) and
use the train set to train the deep learning text generation model.
Once the model is trained we generate a fixed number of passwords (in our experiments up to $10^9$).
We are interested in the number (both total and as ratio) of generated passwords that appear in the test set.

For the first evaluation we train a model on a train dataset and test on the corresponding test dataset, which comes from the same original distribution of passwords. In practice passwords of interest rarely follow a known distribution one could train a model on.
We therefore evaluate our performance by training on the train split of one dataset and evaluating on the
test split of another dataset.

Table (\ref{tab:dl-models-results}) presents all the deep models trained on the `rockyou' or `pwnd' dataset and evaluating on the full `linkedin', `myspace', `yahoo', `seclist', `skullsecurity', `youku' and `zomato' dataset. In the case of `rockyou' and `pwnd' only the test set is used for evaluation. Overall, it is clear from the table that the different training/architecture has a drastic performance impact.

First, focusing on the `rockyou' trained models in Table (\ref{tab:dl-models-results}) we see that the PassGAN benchmark is significantly outperformed by VAE and transformer-based models (up to almost three times on e.g. `rockyou' and `linkedin' datasets). On one hand, this demonstrates how effective model selection may lead to substantial matching improvements; on the other hand, one might speculate that the direct application of GAN-based methods is sub-optimal when one works with password datasets with richer latent structure. In contrast, latent models and more sophisticated representation (AEs and attention) techniques seem much more capable of effectively learning a complex latent structure - for a related analysis of more sophisticated GAN-applications we refer to \cite{Pasquini2019}. Further, the table provides evidence that the top performing `rockyou'-trained models (across most datasets) are our proposed VAE-architecture along with the GPT transformers, where the margin with the second and third competitors (WAE and PassGAN, respectively) is significant. 

In terms of the VAE, a couple of important observations are: 1. Interestingly, although having a very different structure, the VAE performs very similarly to the proposed GPT-models, thus suggesting that perhaps some internal password dataset features (e.g. complexity/margins/topology) are crucially guiding the performance of both approaches; 2. Compared to other autoencoding approaches, such as the WAE, it is clear that VAE could perform almost twice as good in terms of matching. This suggests that the WAE, although aiming to provide better reconstruction and training performance, appears to lack the needed flexibility to adapt an efficient latent space representation by forcing the aggregated posteriors to match the fixed prior distribution \cite{tolstikhin2017wasserstein}.

A similar analysis of the `pwnd' trained models in Table (\ref{tab:dl-models-results}) can be brought forward, where the overall model performance is reduced in comparison to the `rockyou' trained models. Here, in contrast, the transformers seem to have a clear advantage over the autoencoding methods, with VAE and WAE being, respectively, second and third in performance. These observations illustrate the effect of the training dataset's generalization ability - the `rockyou' training appears much richer than the `pwnd' one and, curiously, renders VAE almost equivalent in performance to the attention-driven solutions. 

Note however the effect of training dataset and pretraining procedure on the GPT2-based models.
We denote with GPT2F models that were first trained on a large corpus of regular english text and then only finetuned on password data,
and denote with GPT2S the corresponding model that was trained from scratch only on password data.
We here see that pretraining and finetuning seems to improve the performance on almost all datasets when finetuning happens on the `rockyou' dataset, for some test datasets we see an increase in multiple percentage points over the model trained from scratch.
The opposite effect is happening on the models trained on `pwnd'.
Here we observe the model trained from scratch beating the finetuned model, athough by a smaller margin.
Keeping in mind that the `rockyou' training dataset is significantly smaller than the `pwnd' dataset (10.1M and 219.6M passwords respectively) we hypothesize that finetuning provides additional information on the general use of language and additional vocabulary that may not be present in smaller dataset but available in the much larger `pwnd' dataset.

In Table (\ref{tab:full-results}) we present a thorough description of the performances of all the models proposed/tested in this paper. The models GPT2F, GPT2S, VAE and WAE are trained on both `rockyou' and `pwnd' separately, with exception to the PassGAN that has only been trained on the `rockyou' dataset. 

The results in Table (\ref{tab:full-results}), on one hand, provide a thorough overview of matching-performance on `rockyou' and `pwnd' in terms of the (guessing) sample size. Moreover, we also illustrate the generation of unique passwords in terms of the sample size, where we emphasize the variability in the model's output. Surprisingly, the top performing solutions in this regard are clearly the autoencoding-based ones (WAE and VAE) trained on `pwnd' - although falling behind the transformers in terms of matching, the autoencoders seem to provide much greater output variability. Intuitively, the latent space obtained distributions used for generation have a slower decay, whereas the transformers and GANs correspond to much more concentrated sampling distributions.

Finally, in Table (\ref{tab:segmets-vs-chars-results}) we compared the two different pre-processing methods: character and segment based. Lets denote as VAE-char/WAE-char the models trained on the character based pre-processed `rockyou' dataset and VAE-segment/WAE-segment the models trained on the on the segment based pre-processed `rockyou' dataset. All the models are evaluated on the `linkedin' dataset containing ~47.9M passwords. One can see from the table that models trained on the character based pre-processed dataset generate larger number of unique passwords. The VAE-char is performing better than the VAE-segment however, the WAE-char performs significantly worse than the WAE-segment model.

\begin{table}[!t]
    \centering
    \begin{tabular}{l|crrrrr}
    &&&&&& \\
        Model &
        \begin{tabular}[c]{@{}l@{}}Passwords\\Generated\end{tabular} &
        \begin{tabular}[c]{@{}l@{}} Unique\\Passwords \end{tabular} &
        \multicolumn{2}{c}{\begin{tabular}[c]{@{}l@{}} Matched\\on `rockyou'\end{tabular}} &
        \multicolumn{2}{c}{\begin{tabular}[c]{@{}l@{}} Matched\\on `pwnd' \end{tabular}} \\

        \toprule
        \multirow{4}{*}{GPT2F (rockyou)}
        & $10^6$ & $ 9.96 \times 10^5$ &  32 156 & 1.23\%    &  76 423 & 0.14\%                  \\
        & $10^7$ & $ 8.74 \times 10^6$ &  220 267 & 8.42\%   &  611 137 & 1.11\%                   \\
        & $10^8$ & $ 6.78 \times 10^7$ &  692 514 & 26.5\%   & 2 981 523 & 5.43\%     \\
        & $10^9$ & $ 4.57 \times 10^8$ & 1 140 201 & 43.8\%  & 8 050 063 & 14.65\%     \\ 
        \midrule
        
        \multirow{4}{*}{GPT2S (rockyou)}
        & $10^6$ & $9.47 \times 10^5$ &  33 952 & 1.3\%   &  76 160 & 0.14\% \\
        & $10^7$ & $8.48 \times 10^6$ &  212 494 & 8.13\%   &  578 440 & 1.05\%  \\
        & $10^8$ & $6.62 \times 10^7$ &  637 188 & 24.4\%   & 2 794 198 & 5.08\%  \\
        & $10^9$ & $4.54 \times 10^8$ & 1 090 720 & 41.7\%   & 7 629 127 & 13.9\%  \\

        \toprule
        \multirow{4}{*}{GPT2F (pwnd)}
        & $10^6$ & $ 9.65 \times 10^5$ & 11 130 & 0.43\%   &  42 942 & 0.08\%                   \\
        & $10^7$ & $ 9.23 \times 10^6$ & 71 727 & 2.74\%   &  359 474 & 0.65\%                   \\
        & $10^8$ & $ 8.52 \times 10^7$ & 359 929 & 13.8\%   & 2 662 849 & 4.85\%     \\
        & $10^9$ & $ 6.76 \times 10^8$ & 952 613 & 36.4\%   & 12 155 484 & 22.1\%     \\ 
        \midrule
        
        \multirow{4}{*}{GPT2S (pwnd)}
        & $10^6$ & $ 9.68 \times 10^5$ & 11 207 & 0.43\%   &  44 911 & 0.08\%                   \\
        & $10^7$ & $ 9.27 \times 10^6$ & 74 756 & 2.86\%   &  376 451 & 0.68\%                   \\
        & $10^8$ & $ 8.54 \times 10^7$ & 381 804 & 14.6\%  & 2 798 394 & 5.09\%     \\
        & $10^9$ & $ 6.71 \times 10^8$ & 984 225 & 37.64\%  & 12 494 652 & 22.7\%     \\

        \midrule
        \multirow{4}{*}{PassGAN (rockyou)}
        & $10^6$ & $8.50 \times 10^5$  &  19 835 & 0.76\%   &    47 978 & 0.09\%   \\
        & $10^7$ & $6.84 \times 10^6$  &  79 731 & 3.05\%   &   257 882 & 0.47\%   \\
        & $10^8$ & $4.83 \times 10^7$  & 212 238 & 8.12\%   &   962 761 & 1.75\%   \\
        & $10^9$ & $2.95 \times 10^8$  & 415 859 & 15.9\%   & 2 572 591 & 4.68\%     \\

        \midrule
        \multirow{4}{*}{VAE (rockyou)}
        & $10^6$ & $9.93 \times 10^5$ & 26 814 & 1.03\% & 63 381 & 0.12\%  \\
        & $10^7$ & $9.33 \times 10^6$ & 190 371 & 7.28\% & 519 099 & 0.94\%  \\
        & $10^8$ & $7.86 \times 10^7$ & 638 790 & 24.4\% & 2 782 128 & 5.06\%  \\
        & $10^9$ & $ 5.99 \times 10^8$ & 1 175 420 & 44.9\%  & 8 456 376  & 15.4\%     \\
        
        \midrule
        \multirow{4}{*}{VAE (pwnd)}
        & $10^6$ & $9.99 \times 10^5$   &   2 362 &  0.09\%   &    15 093 & 0.03\%  \\
        & $10^7$ & $9.98 \times 10^6$   &  23 252 &  0.89\%   &   150 699 & 0.27\%  \\
        & $10^8$ & $9.82 \times 10^7$   & 183 202 &  7.01\%   & 1 360 894 & 2.48\%  \\
        & $10^9$ & $8.82 \times 10^8$   & 698 605 & 26.7\%    & 8 020 259 & 14.6\%     \\
        
        \midrule
        \multirow{4}{*}{WAE (rockyou)}
        & $10^6$ & $7.41 \times 10^5$   & 7 666   & 0.29\%   & 21 217 & 0.04\%  \\
        & $10^7$ & $7.14 \times 10^6$   & 56 903  & 2.18\%   & 176 971 & 0.32\%  \\
        & $10^8$ & $6.48 \times 10^7$   & 271 908 & 10.4\%   & 1 106 844 & 2.01\%  \\
        & $10^9$ & $6.62 \times 10^8$   & 738 685 & 28.2\%   & 4 736 310 & 8.62\%     \\
 
        \midrule
        \multirow{4}{*}{WAE (pwnd)}
        & $10^6$ & $7.55 \times 10^5$   & 690     & 0.03\%   & 3 329 & 0.01\%  \\
        & $10^7$ & $7.55 \times 10^6$   & 6 476   & 0.25\%   & 32 318 & 0.06\%  \\
        & $10^8$ & $7.48 \times 10^7$   & 54 145  & 2.07\%   & 280 953 & 0.51\%  \\
        & $10^9$ & $9.50 \times 10^8$   & 306 834 & 11.7\%   & 1 960 948 & 3.57\%     \\
    \end{tabular}
    \caption{Full evaluation for all trained models.
    All rockyou/pwnd models trained on the rockyou/pwnd training dataset (80\% of the corpus)
    and evaluated on the test dataset (20\% of the corpus).}
	\label{tab:full-results}
\end{table}

\begin{table}[t]
    \centering
    \begin{tabular}{l|c rr rr}
    \multicolumn{1}{l}{\textbf{Preprocessing}} &              & \multicolumn{2}{c}{\textbf{Chars}}         & \multicolumn{2}{c}{\textbf{Segment}}  \\  
    \multicolumn{1}{l}{\textbf{Model}}  & \multicolumn{1}{c}{\begin{tabular}[c]{@{}l@{}}Passwords\\Generated\end{tabular}}          & \multicolumn{1}{c}{\begin{tabular}[c]{@{}l@{}} Unique\\Passwords \end{tabular}} & \multicolumn{1}{c}{\begin{tabular}[c]{@{}l@{}} Matched\\Passwords \end{tabular}} & \multicolumn{1}{c}{\begin{tabular}[c]{@{}l@{}} Unique\\Passwords \end{tabular}} & \multicolumn{1}{c}{\begin{tabular}[c]{@{}l@{}} Matched\\Passwords \end{tabular}} \\  \cline{1-6} 
 
        \multirow{4}{*}{VAE}
 
        & $10^6$ &  $9.93 \times 10^5$ & 0.22\%   & $9.53 \times 10^5$ & 0.20\%     \\
        & $10^7$ &  $9.33 \times 10^6$ & 1.80\%   & $8.95 \times 10^6$ & 1.64\%     \\
        & $10^8$ &  $7.86 \times 10^7$ & 9.26\%   & $7.42 \times 10^7$ & 8.01\%     \\
        & $10^9$ &  $5.99 \times 10^8$ & 24.5\%   & $5.74 \times 10^8$ & 21.4\%     \\   
        \midrule
        
        \multirow{4}{*}{WAE}
 
        & $10^6$ &  $9.82 \times 10^5$ & 0.01\%   & $7.41 \times 10^5$ & 0.06\%     \\
        & $10^7$ &  $9.71 \times 10^6$ & 0.06\%   & $7.14 \times 10^6$ & 0.54\%     \\
        & $10^8$ &  $9.52 \times 10^7$ & 0.57\%   & $6.48 \times 10^7$ & 3.51\%     \\
        & $10^9$ &  $9.18 \times 10^8$ & 3.30\%   & $6.62 \times 10^8$ & 14.3\%     \\   

    \end{tabular}
    \caption{Comparing the performance of the models trained on the `rockyou' dataset using the \textit{character} based or \textit{segment} based pre-processing. Models are evaluated on the `linkedin' dataset (47.3M passwords).}
	\label{tab:segmets-vs-chars-results}
\end{table}

\subsection{Application of password augmentation rules}

As previously mentioned, password recovery tool Hashcat does not only offer efficient hashing of password candidates, but also allows for the application of rules,
which can be used to augment and extend a given password list.
A rule is an instruction for augmentation of a single password and is applied to each wordlist entry before hashing.
For some sample rules see Table (\ref{table: hashcat rules}).

\begin{table}[h!]
    \centering
    \begin{tabular}{r|l|l}
    && \\
        rule & description & example \\
        \midrule
        \texttt{l} & lower case string & \texttt{passWord} $\rightarrow$ \texttt{password} \\
        \texttt{c} & capitalize string & \texttt{passWord} $\rightarrow$ \texttt{PassWord} \\
        \texttt{\$1} & append character & \texttt{passWord} $\rightarrow$ \texttt{passWord1} \\
        \texttt{sa@} & switch characters & \texttt{passWord} $\rightarrow$ \texttt{p@ssWord} \\
        \texttt{c sa@ \$1 \$2} & combined rules & \texttt{passWord} $\rightarrow$ \texttt{P@ssWord12} \\
        \texttt{l so0 sa@ ss5} & combined rules & \texttt{passWord} $\rightarrow$ \texttt{p@55w0rd} \\ && \\
    \end{tabular}
    \caption{Sample rules for Hashcat. The tool allows basic string manipulations and advanced transformations by chaining rules.}
    \label{table: hashcat rules}
\end{table}

\begin{table}[h!]
    \centering
    \begin{tabular}{l@{\hskip 0.3cm}r@{\hskip 0.3cm}r@{\hskip 0.3cm}r@{\hskip 0.3cm}r@{\hskip 0.3cm}r}
       model              & \multicolumn{1}{c}{$N_0$}          & \multicolumn{1}{c}{$N^*_0$}            & \multicolumn{1}{c}{$M_0$}               & \multicolumn{1}{c}{$N_\text{best64}$}    & \multicolumn{1}{c}{$M_\text{best64}$} \\ 
       \midrule
        VAE               & $10^8$             & $7.86 \times 10^7$ &  $6.38 \times 10^5$ & $5.18 \times 10^9$   & $12.9 \times 10^5$ \\ 
        WAE               & $10^8$             & $6.48 \times 10^7$ &  $2.72 \times 10^5$ & $4.27 \times 10^9$   & $9.6 \times 10^5$ \\ 
        GPT2S             & $10^8$             & $6.62 \times 10^7$ &  $6.37 \times 10^5$ & $4.36 \times 10^9$   & $11.2 \times 10^5$ \\ 
        GPT2F             & $10^8$             & $6.78 \times 10^7$ &  $6.92 \times 10^5$ & $4.47 \times 10^9$   & $11.0 \times 10^5$ \\
        PassGAN           & $10^8$             & $4.83 \times 10^7$ &  $2.12 \times 10^5$ & $3.18 \times 10^9$   & $7.0 \times 10^5$ \\ 
        `rockyou' (train)   & $1.05 \times 10^7$ & $1.05 \times 10^7$ &  0                & $6.90 \times 10^8$   & $9.1 \times 10^5$ \\
        
        \bottomrule \\
    \end{tabular}
    \caption{Effects of application of Hashcat rules to $10^8$ generated passwords from various models trained on `rockyou' training data split when matching the `rockyou' test split. For comparison we apply rules directly to the training dataset.
    $N_0$: Number of generated passwords or size of dataset.
    $N^*_0$: Number of unique generated passwords or items in the dataset.
    $M_0$: Original matches in target dataset.
    $N_\text{best64}$: Number of password candidates when applying \emph{best64} rules to the unique passwords.
    $M_\text{best64}$: Number of matches in target dataset candidates when applying \emph{best64} rules.
    }
    \label{table: results hashcat}
\end{table}

Application of rules greatly increases the number of password candidates to test and extends our method in a reasonable way:
the deep-learning model generates base password candidates and further rules make sure that slight permutations of the base candidate are also considered.

Table (\ref{table: results hashcat}) shows the effect of application of the rule lists \emph{best64}\footnote{\url{https://github.com/hashcat/hashcat/tree/master/rules/best64.rule}} with 64 rules on our generated password lists.
We generate $10^8$ password with all models trained on the `rockyou' training split and observe the number of matches in the test split before and after application of rules.
Comparing $M_0$ (matches before application of rules) and $M_\text{best64}$ (matches after application of rules) we observe a significant increase in matches for all models.
For comparison we also apply the same ruleset to the `rockyou' training split itself, which generates $0$ matches before application of rules (there are no duplicates in the `rockyou' dataset) and $9.1\times 10^5$ matches with rules.
The training data as basis therefore produces less matches than all models except for the PassGAN.
We conclude that the further application of rules to the output of our models is a reasonable strategy to further increase password recovery performance,
and that training deep-learning models on the training data split for password generation provides value surpassing the application of predefined rules.

\begin{table}[!h]
\renewcommand{\arraystretch}{1.3}
\small
    \centering
    \begin{tabular}{l|rr}
        Model & Unique Passwords & \quad Matches  \\
        \toprule
        
      
      \footnotesize{3-gram Markov Model} &  $4.35 \times 10^8$  &  $4.27 \times 10^6$  \\ %
      \footnotesize{Hashcat -- best64}   &  $6.66 \times 10^8$  &  $7.26 \times 10^6$  \\ %
      \footnotesize{Hashcat -- gen2}     &  $8.49 \times 10^8$  &  $2.55 \times 10^6$  \\ %
      \footnotesize{PCFG v4.1}           &  $9.71 \times 10^8$  &  $12.52 \times 10^6$  \\ %
      \footnotesize{PRINCE v0.22}        &  $9.99 \times 10^8$  &  $1.65 \times 10^6$ \\
      
    
      \midrule
      \footnotesize{PassGAN (ours)} &  $2.95 \times 10^8$  &  $3.2 \times 10^6$  \\ %
      \footnotesize{GPT2S}          &  $4.54 \times 10^8$  &  $8.85 \times 10^6$  \\ %
      \footnotesize{GPT2F}          &  $4.57 \times 10^8$  &  $9.60 \times 10^6$  \\ %
      \footnotesize{VAE}            &  $5.99 \times 10^8$  &  $10.3 \times 10^6$  \\ %
      \footnotesize{VAE\_S}         &  $5.74 \times 10^8$  &  $9.06 \times 10^6$  \\ %
      \footnotesize{WAE}            &  $9.18 \times 10^8$  &  $1.35 \times 10^6$  \\ %
      \footnotesize{WAE\_S}         &  $6.62 \times 10^8$  &  $6.10 \times 10^6$  \\ %
        
    \end{tabular}
    \caption{Results of our evaluation on the `linkedin' dataset (47.3M passwords). All our models were trained on `rockyou' and generated $10^9$ passwords, all models above generated $10^9$ passwords or the maximum number of possible combinations from the `rockyou' training split.}
    \label{tab:linkedin_comparison}
\end{table}

\subsection{Comparison to the established methods}

In order to compare our models to the established methods from Section \ref{sec:related_work},
we evaluate on a third dataset.
We use our training split of `rockyou' (80\%, 10.4M passwords) to generate new passwords using various established methods and evaluate on a subset of the `linkedin' dataset (originally 60.7M passwords).
We prepare this dataset by removing all entries longer than 12, all entries containing non-ascii characters and all entries that also appear in our `rockyou' training split. 
We are left with a test set of 47.3M passwords.

For comparison, we train PCFG\footnote{\url{https://github.com/lakiw/pcfg_cracker}} on a non-unique version of the training split, i.e. passwords appear multiple times in the frequency of the original leak, and generate $10^9$ passwords.
We use Hashcat to apply two rulesets to the training split of unique passwords. 
Ruleset \emph{best64} contains 64 rules and generates $6.9 \times 10^8$ passwords in total.
Ruleset \emph{generated2}\footnote{\url{https://github.com/hashcat/hashcat/tree/master/rules/generated2.rule}} contains 65k rules and generates an exceedingly large number of password candidates, of which we sample $10^9$ passwords.
Both lists are the result of large-scale quantitative evaluations of the effect of various hand-written and machine generated rules on multiple wordlists, password datasets and target hashes.
We additionally train a simple 3-gram Markov Model\footnote{\url{https://github.com/brannondorsey/markov-passwords}} on the unique training split and generate $10^9$ passwords.
Finally we use the PRINCE algorithm\footnote{\url{https://github.com/hashcat/princeprocessor}} to construct $10^9$ passwords of length 4 to 12 from the `rockyou' training set.


For our models, trained on the `rockyou' training split we generate $10^9$ passwords each and count the matches in the `linkedin' test data.
Table (\ref{tab:linkedin_comparison}) shows the results.

We first observe that all trained models recover a significant amount of passwords from the `linkedin' test data.
Ranging from 1.35M (WAE) to 10.3M (VAE) there is large variance in the performance of the individual models.
Interestingly we observe no correlation between number of unique generated passwords and number of matches both in our model and the comparison methods, 
the trained models with most and least unique passwords (2.95M for PassGAN and 9.18M for WAE\_S) match the least number of passwords (3.2M and 1.35M respectively)

Both implementations of VAE and GPT2 respectively achieve very high matching results,
with the character-based VAE representing the top-performer with 10.3M matches.
Only the probabilistic PCFG algorithm can surpass this model by another 20\%.
These trained models additionally score higher than all other comparison methods.

\subsection{Operations in latent space}

\textbf{Similarity in latent space.} The learned latent space by the encoder imposes  geometric connections among latent points that have some semantic similarity in the data space. This means that similar points in data space have latent representation that are close to each other. This property can be used also for password generation. Let us assume that we have the password \verb|veronica2296| and we want to generate variants of this passwords. To this end we encode the password \verb|veronica2296| into its latent representation $\mathbf{z}_t$. We parametrize the posterior using the $\mathbf{z}_t$ as mean ($\boldsymbol{\mu}=\mathbf{z}_t$), next we sample latent codes from that region ($\sigma=0.001$) and we generate passwords $\mathbf{z}_i\sim\mathcal{N}(\boldsymbol{\mu}, \sigma\mathbf{I}$). The results from this task are presented in Table (\ref{tab:latent_space_similarity}). One can see that most of the passwords generated from this region contain the word \verb|veronica| in combination with different number or variants of the name \verb|veronica| ex. \verb|veronico| and a number. This shows that our models have learned semantically meaningful latent space given the training set.

\begin{table}[h]
\centering
\begin{tabular}{lllll}
\verb|veronica2286|  & \verb|veronica296|   & \verb|verogani2297|  & \verb|veronica2229| & \verb|veroicata22U| \\
\verb|veronica22U6|  & \verb|veronic22259|  & \verb|veroinca2297|  & \verb|verolica2296| & \verb|verotEic2246| \\
\verb|verogama2296|  & \verb|veronica2269|  & \verb|veronica22_9| & \verb|veronica239|  & \verb|verolica2298| \\
\verb|veroga_a2986| & \verb|veronic2205|   & \verb|veronaxa2269|  & \verb|veronico2259| & \verb|verolicat269| \\
\verb|verosgaj2!98|  & \verb|vertinac2219|  & \verb|veronica249|   & \verb|verincia22T6| & \verb|veronema2205| \\
\verb|veroina22_6|  & \verb|verozica22_|  & \verb|veroszi2246|   & \verb|veronkea2295| & \verb|verolica2295| \\
\verb|veroneza2269|  & \verb|veron_ma2295| & \verb|verogaxa2299|  & \verb|veronica4298| & \verb|verote5a229D| \\
\verb|veronica20g6|  & \verb|verogala285|   & \verb|verozoca24_7| & \verb|veronic22#6| & \verb|veronoca28/5| \\
\verb|veronico23D|   & \verb|veronema2276|  & \verb|veronica2296|  & \verb|veronican298| & \verb|veronicas25|  \\
\verb|vroricak229|   & \verb|verongic2279|  & \verb|vegaroaka269|  & \verb|veronica2896| & \verb|vetRonic225|  \\ \\
\end{tabular}
\caption{Samples with latent representation close to \textit{veronica2296} latent representation.}
\label{tab:latent_space_similarity}
\end{table}

\textbf{Conditional password generation.} Having latent representation for each password allows us to also do conditional generation. Let \textit{***love***} be a template password. The `*' symbol it is a placeholder for any character defined in the vocabulary. We can condition our model to generate passwords that contain the word `love' in the middle, with three random characters as prefix and suffix. In Table (\ref{tab:conditional_generation}) are presented some conditionally generated samples. For a further thorough analysis of conditional password sampling in terms of EM-based algorithms we refer to \cite{Pasquini2019}. In particular, in this regard our VAE model exhibits a competitive performance as seen in Table (\ref{tab:conditional_generation}).

\begin{table}[h]
\centering
\begin{tabular}{llll}
\verb|9alolove71u| & \verb|nublove85/9| & \verb|miblovenv11| & \verb|cetlovesder| \\
\verb|licloverrs9| & \verb|siclove00me| & \verb|riglover2k|  & \verb|biolove121|  \\
\verb|hicloven3ke| & \verb|failoveye4|  & \verb|n2ulovemswo| & \verb|inudlove12|  \\
\verb|lyaloveji8|  & \verb|gemloveso1|  & \verb|irolovesor|  & \verb|Vealover.v|  \\
\verb|ltelovejr*|  & \verb|vatlover10|  & \verb|mejlovey4u|  & \verb|sewalover79| \\ \\
\end{tabular}
\caption{Conditional generation of passwords. We condition the generation on \textit{***love***}.}
\label{tab:conditional_generation}
\end{table}

\textbf{Interpolation between two passwords.} Another possibility with the VAE models is to do interpolation between two passwords. Let us assume we have the password $A$ and the password $B$. We use the encoder to obtain the latent representations $\mathbf{z}_A$ and $\mathbf{z}_B$ for $A$ and $B$ respectively. Next, we sample latent representations from the linear path former by $\mathbf{z}_A$ and $\mathbf{z}_B$ and using the decoder we obtain the corresponding passwords. The process of sampling new passwords in this way is called \textit{interpolation}. In Table (\ref{tab:interpolations}) are given some example interpolations. Each column in the table is an example interpolation. The top row is the password $A$ and the bottom row is the password $B$. In the middle are the passwords generated from the latent representations sampled form the linear path between $A$ and $B$.

\begin{table}[h]
\centering

\begin{tabular}{llllll}

\verb|remington223| & \verb|pepegrillo16|  & \verb|newthanakorn|   & \verb|musikALE0991| & \verb|DIMAYUGAluna| & \verb|fressikarosa| \\
\verb|rtninton2131| & \verb|popegrillo16|  & \verb|nevwhanfaror|   & \verb|muwikO134902| & \verb|DIMAYUYAUMEM| & \verb|krezsickiess| \\
\verb|Atnitton2133| & \verb|petckrllole7|  & \verb|new_hanaqira|   & \verb|muwIE1250894| & \verb|DIMuYuYnuMAY| & \verb|kryesikarish| \\
\verb|ntn123to2131| & \verb|hoperillel62|  & \verb|newsafarhrig|   & \verb|memIE1284989| & \verb|GIMAYuYuuakH| & \verb|kynesiha2308| \\
\verb|rmnm2et21213| & \verb|agterillo652|  & \verb|18stahankoua|   & \verb|mTO130281948| & \verb|VUHuwuNAuMMe| & \verb|kynayiha2s09| \\
\verb|nnnmot121032| & \verb|ktanlilel657|  & \verb|18s7tafabano|   & \verb|K72636481934| & \verb|WHfluvusue85| & \verb|cynayoshi_09| \\
\verb|nnd231523465| & \verb|allininal505|  & \verb|129swokfaton|   & \verb|172639451658| & \verb|Tofluousuck9| & \verb|cy2ayueh2990| \\
\verb|ontit1321642| & \verb|allinllang08|  & \verb|1287tockokno|   & \verb|172639293483| & \verb|WofdmousuS86| & \verb|clyayosh1090| \\
\verb|andiew123456| & \verb|alliannan680|  & \verb|12960ocktint|   & \verb|172839481614| & \verb|moodmous6885| & \verb|clya23081099| \\
\verb|andrew123456| & \verb|allianna0685|  & \verb|1297bucktown|   & \verb|172839456123| & \verb|woodmouse598| & \verb|clya23081990| \\ \\
\end{tabular}
\caption{Example interpolations. Each column is one interpolation where the top row is the start password $A$ and the last row is the end password $B$. In between are generated passwords.}
\label{tab:interpolations}
\end{table}

\subsection{Intersection of generated passwords}
\begin{figure}
     \centering
     \begin{subfigure}[t]{0.32\textwidth}
         \centering
         \includegraphics[width=\textwidth]{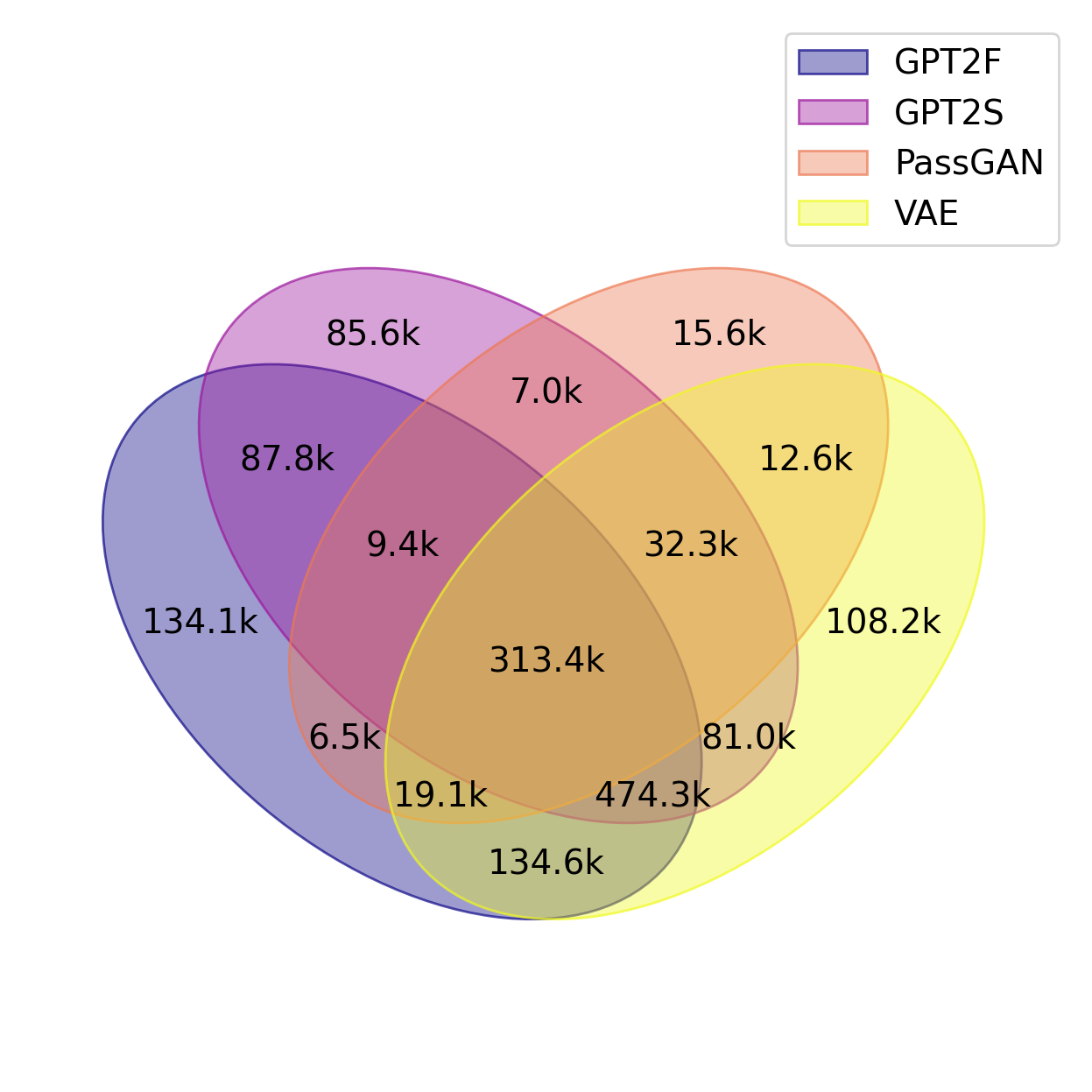}
         \caption{Models trained on `rockyou' matched on `rockyou' test set.}
     \label{fig:venn_rockyou_rockyou}
     \end{subfigure}
     \hfill 
     \begin{subfigure}[t]{0.32\textwidth}
         \centering
         \includegraphics[width=\textwidth]{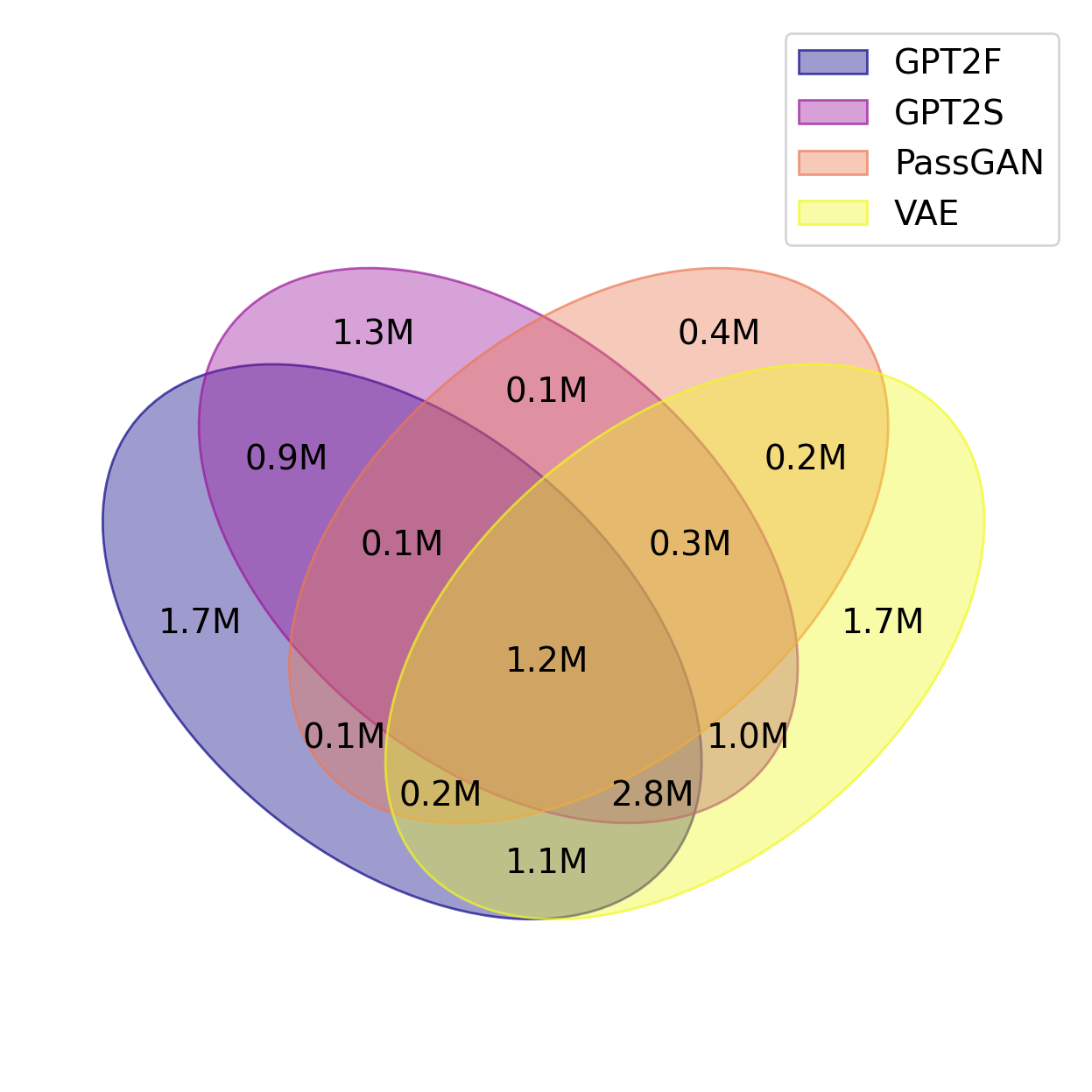}
         \caption{Models trained on `rockyou' matched on `pwnd' test set.}
     \label{fig:venn_rockyou_pwnd}
     \end{subfigure}
     \hfill
     \begin{subfigure}[t]{0.32\textwidth}
         \centering
         \includegraphics[width=\textwidth]{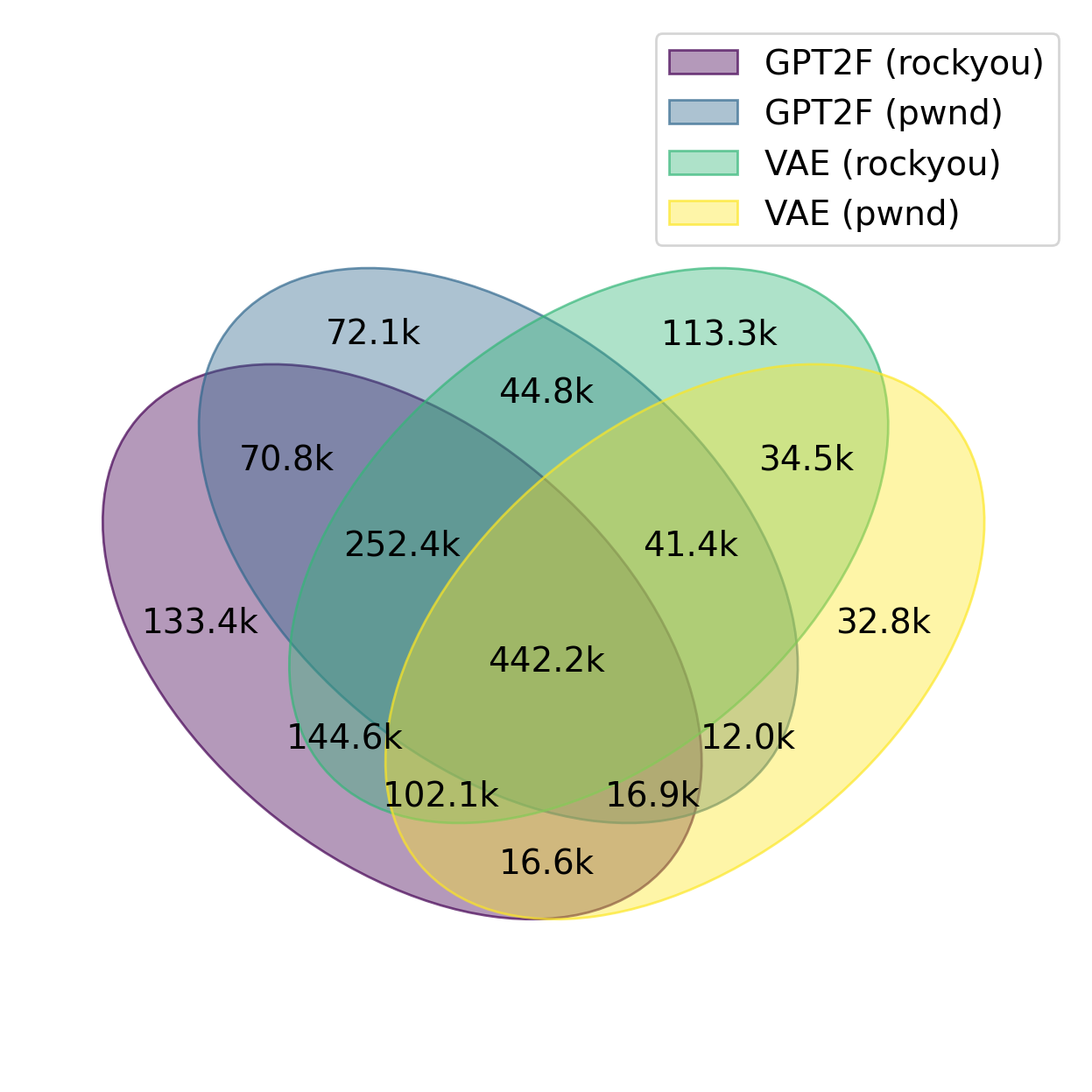}
         \caption{Models trained on `rockyou' or `pwnd' matched on `rockyou' test set.}
     \label{fig:venn_rockyou_pwnd_rockyou}
     \end{subfigure}
     \caption{
     We compare the passwords matched by various models on different datasets to analyse whether the output of different models is sufficiently different. Numbers given in thousands or millions.}
     \label{fig:venn}
\end{figure}

Table (\ref{tab:most_common}) showed that different model architectures qualitatively seem to generate vastly different passwords,
however the question remains if different architectures and different training datasets actually lead to a quantitatively different generation
and match different subsets of the test dataset.
Take for instance the two GPT2-based models trained on the `rockyou' dataset.
Tables (\ref{tab:most_common}) and (\ref{tab:most_common_frequencies}) show qualitatively a different distribution on generated passwords,
however Table (\ref{tab:full-results}) tells us that both models match around $40\%$ of the passwords in the `rockyou' test dataset.
Figure (\ref{fig:venn}) therefore analyses the intersection of passwords matched on the `rockyou' and `pwnd' test dataset between the models.

Figure (\ref{fig:venn_rockyou_rockyou}) shows the intersections of the models GPT2F, GPT2S, PassGAN and VAE, trained on the `rockyou' dataset and matched on the `rockyou' training set.
We see that while there is a large overlap between matched passwords, with 313k passwords appearing in every model output,
each model still has a significant amount of unique passwords.
In this respect there are large differences between the models.
While the GPT2F model recovered 134k password no other model could recover, our PassGAN implementation only had 16k unique found passwords,
and we find the largest intersection in the passwords that all models except PassGAN could recover.
All models in total matched around 1.5M passwords, or 30\% more than the best single model (VAE with 1.18M recovered passwords).

The effect is even more pronounced when evaluating the same models, trained on the `rockyou' dataset, tested on the `pwnd' dataset in Figure (\ref{fig:venn_rockyou_pwnd}).
We again see a similar behaviour as in Figure (\ref{fig:venn_rockyou_rockyou}),
with all models generating a significant amount of unique passwords matched on the test set.
Again the GPT2 models and the VAE generate more unique found passwords than the PassGAN (1.3M -- 1.7M against 0.4M passwords),
and the intersection between the three non-PassGAN models represent the largest subset of found passwords (2.8M passwords).
In total the models generated 13.1M unique passwords found in the `pwnd' data,
or 55\% more than the best single model (VAE with 8.45M recovered passwords).

Finally Figure (\ref{fig:venn_rockyou_pwnd_rockyou}) analyses the effect of training data on the matched passwords.
We compare the GPT2F and VAE models, both trained on `rockyou' and pwnd.
While now the largest subset lies in the intersection of all four models (442k),
again many passwords in the `rockyou' test set are only found by one model.
The models trained on `rockyou' unsurprisingly offer a better performace than the models trained on pwnd.
While the `rockyou' models together recovered 391k passwords that none of the `pwnd' models could generate,
the `pwnd' models together only found 117k unique passwords.
Similar to the analysis of Figure (\ref{fig:venn_rockyou_rockyou}), all models recover a total of 1.5M passwords, or 30\% more than the best single model (VAE (rockyou) with 1.18M recovered passwords).

We conclude that there is a significant overlap in recovered passwords from each model,
which we do not find surprising given that the models share common architectures or training datasets 
and the best models alone already recover over 40\% of the entire test set.
However the diagrams also reveal that many models find a large amount of unique passwords,
and the set of all password generations outperforms any single model by a large margin.
Each model therefore offers unique password recovery functionality and could be considered in a password recovery task.


\section{Conclusion and Future Work}

The present work illustrates various deep learning password generation techniques. Conducting a thorough unified analysis we discuss password-matching capabilities, variability and quality of sampling and robustness in training. On one hand, we bridge and extend previous methods based on attention schemes, GANs and Wasserstein autoencoding; on the other hand, we provide a promising novel approach based on Variational Autoencoders that allows for efficient latent space modeling and further sampling mechanisms. Lastly, we hope our work will facilitate and provide benchmark lines for further deep learning and ML practitioners interested in the field of password guessing.

In terms of further investigation, the application of deep learning techniques to password generation poses further intriguing questions on the interplay between classical probabilistic methods and neural networks, where one would ultimately hope to construct more efficient and reliable domain-inspired password representation schemes - e.g. based on carefully crafted fragmentations. 

\section*{Acknowledgment}
This project was funded by the Federal Ministry of Education and Research (BMBF), FZK: 16KIS0818.
The authors of this work were supported by the Competence Center for Machine Learning Rhine Ruhr (ML2R) which is funded by the Federal Ministry of Education and Research of Germany (grant nos. 01|S18038B, 01|S18038C). We gratefully acknowledge this support.

\bibliographystyle{abbrv}
\bibliography{main}

\end{document}